\documentclass[conference]{IEEEtran}
\IEEEoverridecommandlockouts
\usepackage{cite}
\usepackage{amsmath}
\usepackage{amssymb}
\usepackage{amsfonts}
\usepackage{graphicx}
\usepackage{textcomp}
\usepackage{xcolor}

\usepackage{booktabs}
\usepackage{amsthm}
\usepackage{amsbsy}
\usepackage{algorithm}
\usepackage{algorithmicx}
\usepackage[noend]{algpseudocode}
\usepackage[htt]{hyphenat}
\usepackage{multirow}
\usepackage{enumitem}
\usepackage{float}
\usepackage{bbm}
\PassOptionsToPackage{hyphens}{url}\usepackage{hyperref}

\def\BibTeX{{\rm B\kern-.05em{\sc i\kern-.025em b}\kern-.08em
    T\kern-.1667em\lower.7ex\hbox{E}\kern-.125emX}}

\newcommand*\Input[1]{\Statex \textbf{Input:} #1}
\newcommand*\Output[1]{\Statex \textbf{Output:} #1}
\algrenewcommand\alglinenumber[1]{#1}

\newtheoremstyle{mystyle}
  {\abovedisplayskip}     
  {\belowdisplayskip}     
  {}                      
  {}                      
  {\bfseries}             
  {.}                     
  {.5em}                  
  {}                      

\theoremstyle{mystyle}
\newtheorem{define}{Definition}

\setlist[itemize]{leftmargin=*, topsep=0pt, itemsep=0pt, parsep=0pt, partopsep=0pt}
\setlist[enumerate]{leftmargin=*, topsep=0pt, itemsep=0pt, parsep=0pt, partopsep=0pt}

\begin{document}

\title{Towards a Flexible Embedding Learning Framework}

\author{
\IEEEauthorblockN{Chin-Chia Michael Yeh, Dhruv Gelda, Zhongfang Zhuang, Yan Zheng, Liang Gou, and Wei Zhang}
\IEEEauthorblockA{\textit{Visa Research} \\
\{miyeh, dhgelda, zzhuang, yazheng, ligou, wzhan\}@visa.com}
}

\maketitle

\begin{abstract}
Representation learning is a fundamental building block for analyzing entities in a database.
While the existing embedding learning methods are effective in various data mining problems, their applicability is often limited because these methods have pre-determined assumptions on the type of semantics captured by the learned embeddings, and the assumptions may not well align with specific downstream tasks.
In this work, we propose an embedding learning framework that 1) uses an input format that is agnostic to input data type, 2) is flexible in terms of the relationships that can be embedded into the learned representations, and 3) provides an intuitive pathway to incorporate domain knowledge into the embedding learning process.
Our proposed framework utilizes a set of  \textit{entity-relation-matrices} as the input, which quantifies the affinities among different entities in the database.
Moreover, a sampling mechanism is carefully designed to establish a direct connection between the input and the information captured by the output embeddings.
To complete the representation learning toolbox, we also outline a simple yet effective post-processing technique to properly visualize the learned embeddings.
Our empirical results demonstrate that the proposed framework, in conjunction with a set of relevant entity-relation-matrices, outperforms the existing state-of-the-art approaches in various data mining tasks.

\end{abstract}

\begin{IEEEkeywords}
embedding learning, representation learning, unsupervised learning, visualization, clustering
\end{IEEEkeywords}

\begin{figure*}[t]
    \centering
    \includegraphics[width=0.85\linewidth]{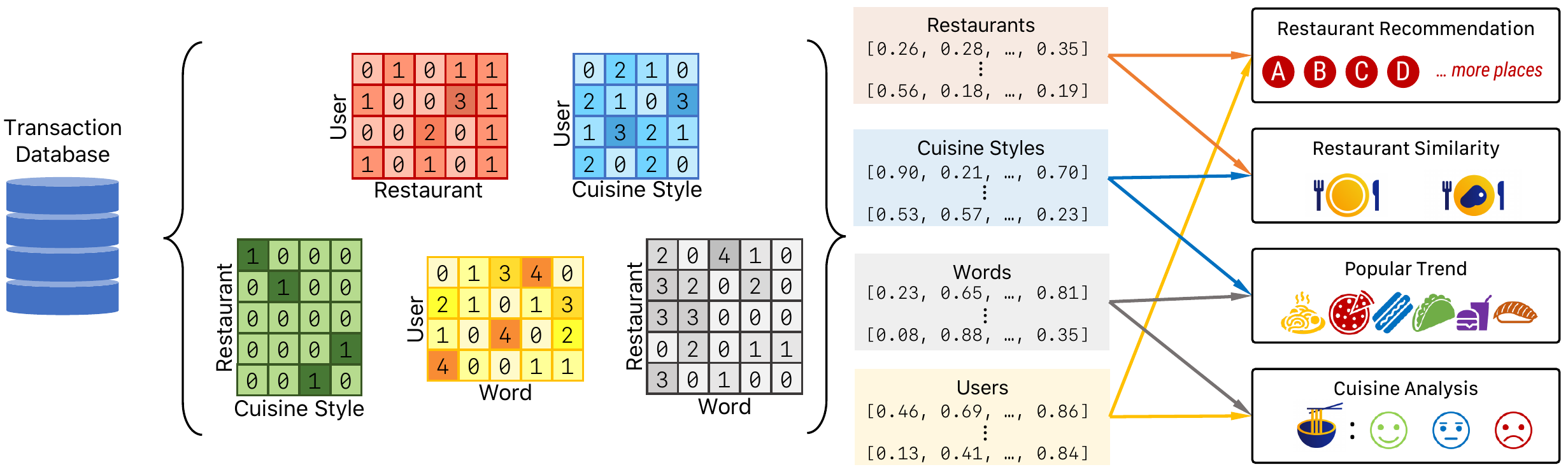}
    \caption{
    Motivating Example.
    A use case of the proposed framework for credit card transaction database.
    Five matrices describing the relationships of restaurant, cuisine style, user, and word appears in name of restaurant are extracted from credit card transaction database.
    Embeddings are learned using all extracted matrices and used by the downstream tasks.
    }
    \label{fig_motivate}
\end{figure*}

\section{Introduction}
\label{intro}
Embedding learning has emerged as a powerful technique for analyzing text~\cite{mikolov2013efficient,le2014distributed,camacho2018word} by encapsulating relevant information into vectors that could be utilized for downstream tasks.



A majority of the existing embedding learning methods implicitly~\cite{levy2014neural,qiu2018network} or explicitly~\cite{pennington2014glove} perform matrix factorization on a matrix that describes the relationship among different \textit{entities} in a given database.
The difference lies in how the matrix is \textit{indirectly} constructed from the data: word2vec~\cite{mikolov2013efficient} factorizes the pointwise mutual information (PMI) matrix derived from \textit{word}-\textit{contextual word} matrix~\cite{levy2014neural,arora2015rand, allen2019analogies}, DeepWalk~\cite{perozzi2014deepwalk} factorizes the PMI matrix derived from the \textit{node}-\textit{contextual node} matrix~\cite{qiu2018network}, and \textit{GloVe}~\cite{pennington2014glove} factorizes the \textit{word}-\textit{contextual word} co-occurrence matrix.
While the aforementioned methods have achieved state-of-the-art performance in various data mining tasks, they focus on learning representations from a single relationship type (e.g., distributional hypothesis in word2vec).

However, the similarity among entities could consist of multiple relationship types.
For example, two restaurants in a credit card transaction database~\cite{du2019pcard} could be related because they share the same set of customers or cuisine style or even if they share similar names (see Fig.~\ref{fig_motivate}).
If only one of the relationship types is considered, the resulting embeddings will capture \textit{incomplete} information about the restaurants.
Therefore, an embedding learning framework needs to learn from a \textit{set of relationship matrices} for better flexibility in capturing heterogeneous relationships.

Another family of embedding learning methods that could solve the representation learning problem on data with the heterogeneous relationship is the knowledge graph-based methods~\cite{nickel2011three,bordes2013translating,wang2017knowledge}.
These methods capture the heterogeneous relationships among entities through learning separate models in addition to the entity embeddings.
That is, not all the input information is captured by the entity embedding, and part of the information would be transferred to the relationship model instead.
Since the goal of representation learning is to learn the embedding for each entity, creating an additional relationship model only adds unnecessary complication to the followup analysis (e.g., clustering or visualization).
Knowledge graph-based methods are flexible but not suitable for learning representation.

To address the limitations of existing methods, we propose an embedding learning framework where the input to the algorithm is a set of matrices, namely \textit{entity-relation-matrices}, each describing a different relationship among the entities in the database.
By using several entity-relation-matrices as the input, the proposed framework has the following properties:
\begin{itemize}
    \item The capability of modeling multiple heterogeneous relationships from the given database.
    \item Flexibility in terms of the possible methods for the construction of input matrices, \textit{any} suitable measure of affinity between the entities constitutes a valid entity-relation-matrix.
    \item The entity-relation-matrix representation is \textit{agnostic} to the input data type.
    \item An intuitive pathway to incorporate domain knowledge into the embedding learning process.
    To encode any desired information into the embedding, we need to focus on designing the relevant matrix.
\end{itemize}

\noindent To effectively use the information presented in entity-relation-matrices, we design the sampling mechanism during the optimization process inspired by the kernel normalization technique in multiple kernel learning~\cite{bucak2013multiple,kloft2011lp}.
In addition to the proposed framework, we also provide a simple post-processing technique to visualize the learned embeddings accurately.
We demonstrate that directly applying off-the-shelf methods like multi-dimensional scaling (MDS)~\cite{kruskal1964multidimensional} or $t$-SNE~\cite{maaten2008visualizing} would fail to produce meaningful visualization of the learned embedding.
We empirically demonstrate that the proposed algorithm, along with relevant relation-matrices, outperforms the existing state-of-the-art embedding approaches in various data mining tasks.

\section{Definitions}
\label{notation}

We first introduce the concepts of \textit{entity} and \textit{entity-set} as they are the basic building blocks of our framework:
\begin{define}[Entity]
    An \textit{entity} $e$ is something that has a unique and independent existence within a given database $D$.
\end{define}

\begin{define}[Entity-Set]
    An \textit{entity-set} $\mathbf{E} = \{e_i|0 \leq i < n\}$ is the set of all entities that exist in a given database $D$, where $n$ is the total number of entities.
\end{define}
For example, in the credit-card transaction database, each \textit{unique user}, \textit{restaurant} or \textit{cuisine style} represents an entity, whereas all the users, restaurants and cuisine styles together form the entity-set.

\begin{define}[Entity-Type]
    An \textit{entity-type} $t$ is a mutually exclusive label categorizing entities based on conceptual similarity.
\end{define}

The credit-card transaction database consists of four entity-types: user-type, restaurant-type, word-type, and cuisine style-type.

\begin{define}[Entity-Type-Set]
    Given a database $D$, an \textit{entity-type-set} $\mathbf{T}$ is defined as the set of all entity-types.
    Formally, $\mathbf{T} = \{t_i|0 \leq i < m\}$, where $m$ is the number of entity-types in $D$.
\end{define}

We now define the inputs to the embedding learning algorithm: \textit{entity-relation-matrix} and \textit{entity-relation-matrix-set}.

\begin{define}[Entity-Relation-Matrix]
    An \textit{entity-relation-matrix} $M \in \mathbb{R}_{\geq 0}^{n_i \times n_j}$ describes the relationship between entities of entity-type $t_i$ and entities of entity-type $t_j$.
    $n_i$ is the number of type-$t_i$ entities, and $n_j$ is the number of type-$t_j$ entities.
\end{define}

\begin{define}[Entity-Relation-Matrix-Set]
    Given a database $D$, an \textit{entity-relation-matrix-set} $\mathbf{M}$ is the set of all entity-relation-matrices.
\end{define}

Finally, the objective of the proposed \textit{embedding learning algorithm} is defined as:

\begin{define}
    Given the entity-set $\mathbf{E}$ and the entity-relation-matrix-set $\mathbf{M}$ of a given database $D$, the embedding framework computes the embedding $v_i \in \mathbb{R}^d$ for each $e_i \in \mathbf{E}$ such that the entity relationship stored in $\mathbf{M}$ is preserved in the $d$-dimensional space.
    Note that $\mathbf{V}$ is used to denote the \textit{embedding-set}, and $v_i \in \mathbf{V}$ is the corresponding embedding for entity $e_i \in \mathbf{E}$.
\end{define}

\begin{table}[t]
\centering
\caption{Summary of notation}
\label{tab_notation}
\resizebox{0.85\columnwidth}{!}{
\begin{tabular}{cl|cl}
\toprule
Symbol        & Description     & Symbol        & Description                \\ \hline \midrule
$D$           & Database        & $M$           & Entity-relation-matrix     \\
$e$           & Entity          & $M[i, j]$     & $ij$th entry of $M$        \\
$e_i$         & $i$th entity    & $\mathbf{M}$  & Entity-relation-matrix-set \\
$\mathbf{E}$  & Entity-set      & $v$           & Embedding                  \\
$t$           & Entity-type     & $v_i$         & Embedding for $e_i$        \\
$\mathbf{T}$  & Entity-type-set & $\mathbf{V}$  & Embedding-set              \\
\bottomrule
\end{tabular}
}
\end{table}



\section{Methodology}
\label{methodology}
We now discuss the algorithm behind the proposed framework in detail (Algorithm~\ref{alg_mulmat}) and explain how it learns embeddings from multiple matrices derived from a database.
The input to the algorithm includes information about the database: entity-relation-matrix-set $\mathbf{M}$ and entity-set $\mathbf{E}$; and also the hyperparameters associated with the learning process: number of iteration $n_\text{iter}$, number of negative samples $n_\text{neg}$ and learning rate $\eta$.

\begin{algorithm}[ht]
    \centering
    \caption{Embedding Learning with Multiple Matrices\label{alg_mulmat}}
    {\footnotesize
    \begin{algorithmic}[1]
        \Input{entity-relation-matrix-set $\mathbf{M}$, entity-set $\mathbf{E}$, number of iteration $n_\text{iter}$, number of negative sample $n_\text{neg}$ and learning rate $\eta$}
        \Output{embedding-set $\mathbf{V}$}
        \Function{SkipGramMulMat}{$\mathbf{M}, \mathbf{E}, n_\text{iter}, n_\text{neg}, \eta$}
        \State $\mathbf{V} \gets \textrm{ InitializeEmbeddings}()$
        \For{$i \gets 0 \textrm{ \textbf{to} } n_\text{iter}$}
        \For{$\textrm{\textbf{each} } M \in \mathbf{M}$}
        \State $e_p, e_q \gets \textrm{ Sampling}(M)$
        \State $t_q \gets \textrm{ GetEntityType}(e_q)$
        \State $\mathbf{E}_\text{neg} \gets \textrm{ NegativeSampling}(\mathbf{E}, t_q, n_\text{neg})$
        \State $\mathbf{V} \gets \textrm{ UpdateEmbedding}(\mathbf{V}, e_p, e_q, \mathbf{E}_\text{neg}, \eta)$
        \EndFor
        \EndFor
        \State \Return{$\mathbf{V}$}
        \EndFunction
    \end{algorithmic}
    }
\end{algorithm}

First, we initialize the embedding for each entity with a random vector, see line 2 of Algorithm~\ref{alg_mulmat}.
Each step of the outer loop (lines 3-9) is an iteration of the learning algorithm, and each step of the inner loop (lines 4-8) is an iteration over an entity-relation-matrix $M$ in $\mathbf{M}$.
Next, we sample a pair of entities, say $e_p$ and $e_q$, based on the corresponding strength of the relationship between them as given by $M$, see line 5.
Since we iterate over all the entity-relation-matrices and sample each matrix independently, the probability of the entity pair associated with $M[i, j]$ being sampled is $P[i,j] = \frac{M[i,j]}{\sum_{i,j}M[i, j]}$.
The weighted sampling scheme can be efficiently implemented using the alias method~\cite{tang2015line,li2014reducing} with $O(1)$ time complexity.
Next, we determine the entity-type of $e_q$, which is stored in $t_q$ (line 6), and perform negative sampling where $n_\text{neg}$ entities of type $t_q$ are drawn uniformly
from $\mathbf{E}$, stored in $\mathbf{E}_\text{neg}$ (line 7).
The embedding-set $\mathbf{V}$ (line 8) is then updated with the sampled entities using gradient descent.
For an update based on a given entity-relation-matrix~$M$, the gradient is computed by differentiating Equation~\ref{eq_loss}.

\begin{equation}
    \label{eq_loss}
    {\footnotesize
    \begin{array}{lll}
        loss(v_p, v_q, \mathbf{V}_\text{neg}) & = & - \log{\frac{1}{1 + exp(-v_p^\intercal v_q)}}                                  \\
                                &   & - \sum_{v_{\text{neg},i} \in \mathbf{V}_\text{neg}}\log{\frac{1}{1 + exp(v_p^\intercal v_{\text{neg},i})}}
    \end{array}
    }
\end{equation}

\noindent Specifically, $v_p$ and $v_q$ are updated according to Equation~\ref{eq_update01}:

\begin{equation}
    \label{eq_update01}
    \left\{
    \begin{array}{lll}
        v_q^\text{new} & \gets & v_q - \eta \left(\frac{1}{1 + exp(-v_p^\intercal v_q)} - 1\right) v_p \\[1pt]
        v_p^\text{new} & \gets & v_p - \eta \left(\frac{1}{1 + exp(-v_p^\intercal v_q)} - 1\right) v_q
    \end{array}
    \right.
\end{equation}

\noindent Likewise, $v_p$ and $v_{\text{neg},i} \in \mathbf{V}_\text{neg}$ are updated according to Equation~\ref{eq_update0neg}:

\begin{equation}
    \label{eq_update0neg}
    \left\{
    \begin{array}{lll}
        v_{\text{neg},i}^\text{new} & \gets & v_{\text{neg},i} - \eta \frac{1}{1 + exp(-v_p^\intercal v_{\text{neg},i})} v_p \\
        v_p^\text{new} & \gets & v_p - \eta \frac{1}{1 + exp(-v_p^\intercal v_{\text{neg},i})} v_{\text{neg},i}
    \end{array}
    \right.
\end{equation}

\noindent According to \cite{levy2014neural,arora2015rand,allen2019analogies}, it has been hypothesized that minimizing the loss function (Equation~\ref{eq_loss}) is akin to implicitly factorizing a matrix that contains the pointwise mutual information (PMI)\footnote{PMI for a given entity pair can be derived from the entity-relation-matrix as $ PMI(e_p,e_q) = \log \frac{P[e_p,e_q]}{\sum_p P[e_p,e_q] \sum_q P[e_p,e_q]}.$} of the entity pair.
The updated embedding-set $\mathbf{V}$ is then returned in line 9.

\noindent \textbf{Why do we sample each $M \in \mathbf{M}$ independently?}

As mentioned before, the sampling probability for each pair of entities $(e_i,e_j)$ using the independent sampling scheme is given by $P[i,j] = \frac{M[i,j]}{\sum_{i,j}M[i, j]}$.
Alternatively, the normalization constant could be global and given by $\sum_{M \in \mathbf{M}} \sum_{i,j}M[i, j]$.
However, such a sampling approach has a severe shortcoming in comparison to the independent sampling scheme: it could completely ignore some of the entity-relation-matrices, especially if the ``units" of raw relationship strength in each matrix are not commensurate and differ significantly in value.

\begin{figure}[ht]
    \centering
    \includegraphics[page=18, width=1\linewidth, trim={3.89in 2.75in 3.89in 2.77in},clip]{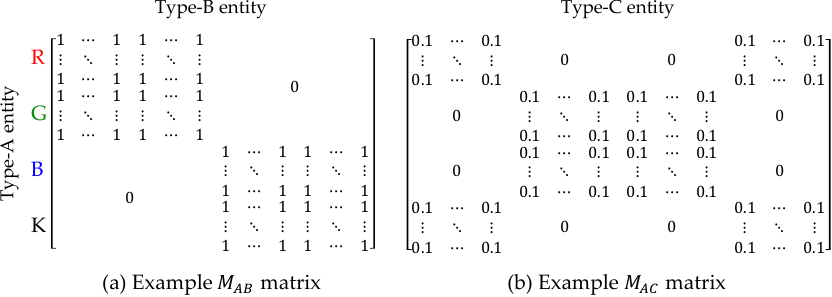}
    \caption{Example entity-relation-matrices.}
    \label{fig_samp_data}
\end{figure}

For instance, consider both $M_{AB}$ and $M_{AC}$ are 20 $\times$ 20 entity-relation-matrix as shown in Fig.~\ref{fig_samp_data}. $M_{AB}$ is constructed by counting the co-occurrences of type-$A$ entities and type-$B$ entities, while $M_{AC}$ contains the co-occurrences of type-$A$ entities and type-$C$ entities normalized by tf-idf.
The diagonal of $M_{AB}$ is occupied by four sub-matrices of size 10 $\times$ 10, each filled with ones, whereas the remaining elements in the matrix are all zeros.
For $M_{AC}$, the center of the matrix is occupied by a 10 $\times$ 10 sub-matrix filled with 0.1, the four corners are each occupied by a 5 $\times$ 5 sub-matrix filled with 0.1, and the rest of the matrix consist of zeros.

\begin{figure}[ht]
    \centering
    \includegraphics[width=0.9\linewidth]{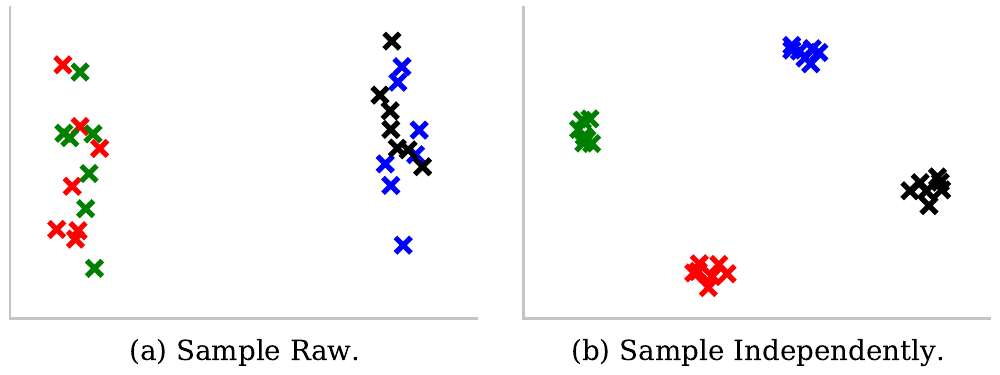}
    \caption{Independent sampling scheme benefit the learning process.}
    \label{fig_sample}
\end{figure}

The independent sampling scheme ensures that both $M_{AB}$ and $M_{AC}$ are utilized equally.
If we naively sample $\mathbf{M}$ with the raw relationship strength, the majority of the entity pairs used for embedding learning (Algorithm~\ref{alg_mulmat}, Line 5) are type-$A$ and type-$B$ entities from $M_{AB}$.
Consequently, the resulting type-$A$ embeddings are unable to capture the information provided by $M_{AC}$, and the learning process fails to separate the four type-$A$ clusters.
Only two clusters (i.e., {\color{red}R}{\color{green}G} cluster and {\color{blue}B}K cluster) are produced as as shown in Fig.~\ref{fig_sample}.a.
However, if we sample $M_{AB}$ and $M_{AC}$ independently, the structure of type-$A$ entities is learned from both $M_{AB}$ and $M_{AC}$ and the type-$A$ entities are clustered into four clusters, i.e., \{{\color{red}R}, {\color{green}G}, {\color{blue}B}, K\} (see Fig.~\ref{fig_sample}.b).

While the independent sampling scheme utilizes \textit{all} matrices in the entity-relation-matrix-set equally, the proposed algorithm could be adapted quite easily to provide preferential weights to the different matrices, if required based on domain knowledge or downstream tasks.
Specifically, the second term in each equation given in Equation~\ref{eq_update01} and~\ref{eq_update0neg} could be modified as

\begin{equation}
    \label{eq_updatevq_new}
    v_q^\text{new} \gets v_q - \eta \alpha \left(\frac{1}{1 + exp(-v_p^\intercal v_q)} - 1\right) v_p,
\end{equation}

\noindent where $\alpha \in [0,1]$ and chosen independently for each matrix.
We use the equation for updating $v_q$ as an example, and the same modification is applied to the other questions in Equation~\ref{eq_update01} and~\ref{eq_update0neg}.

\begin{figure}[ht]
    \centering
    \includegraphics[width=0.99\linewidth]{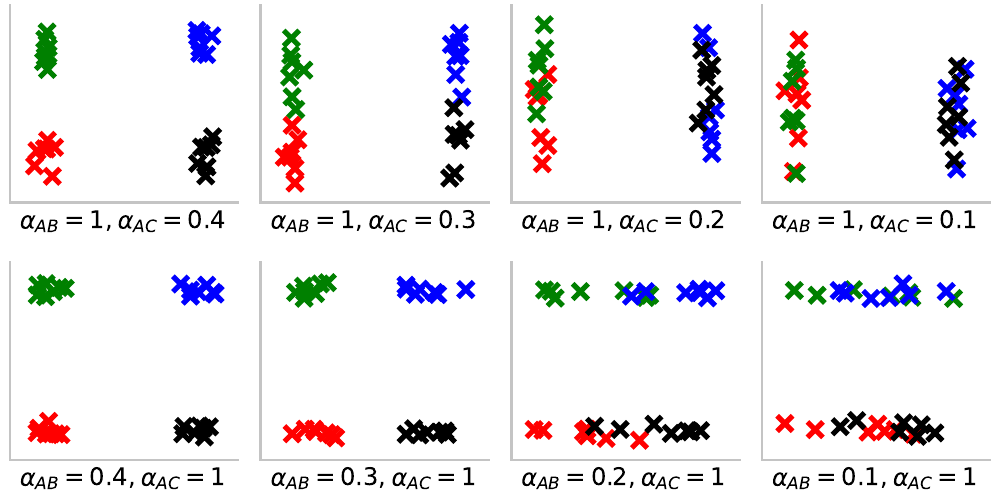}
    \caption{Varying $\alpha_{AB}$ and $\alpha_{AC}$ change the amount of information provided by the corresponding matrices.}
    \label{fig_weight}
\end{figure}

To examine the effect of $\alpha$ on the result embeddings, we learns embeddings using the same $M_{AB}$ and $M_{AC}$ from Fig.~\ref{fig_samp_data} with different settings of $\alpha_{AB}$ and $\alpha_{AC}$.
The result embeddings are shown in Fig.~\ref{fig_weight}.
As expected, the effect of $M_{AB}$ is diminishing as we reduce $\alpha_{AB}$; a similar effect can be observed when $\alpha_{AC}$ is reduced.

The importance of an independent sampling scheme can also be intuitively understood by drawing comparisons with kernel normalization in \textit{multiple kernel learning}.
Each entity-relation-matrix can be considered analogous to a kernel in multiple kernel learning framework as both the entity-relation-matrix and the individual kernel measure some form of affinity between entities/objects.
Normalizing each kernel when applying multiple kernel learning algorithms is essential because it represents an ``uninformative prior" as to which kernels would be more useful~\cite{bucak2013multiple,kloft2011lp}.
Similarly, the independent sampling scheme balances the relative contribution of different entity-relation-matrices to the embedding learning process by treating each matrix equally.
On top of that, the relative contribution of different matrices can be further adjusted by varying $\alpha$ associated with each matrix.

\noindent \textbf{How do we prepare entity-relation-matrices?}

\begin{figure}[ht]
    \centering
    \includegraphics[page=19, width=0.85\linewidth, trim={4.50in 1.14in 4.51in 2.06in},clip]{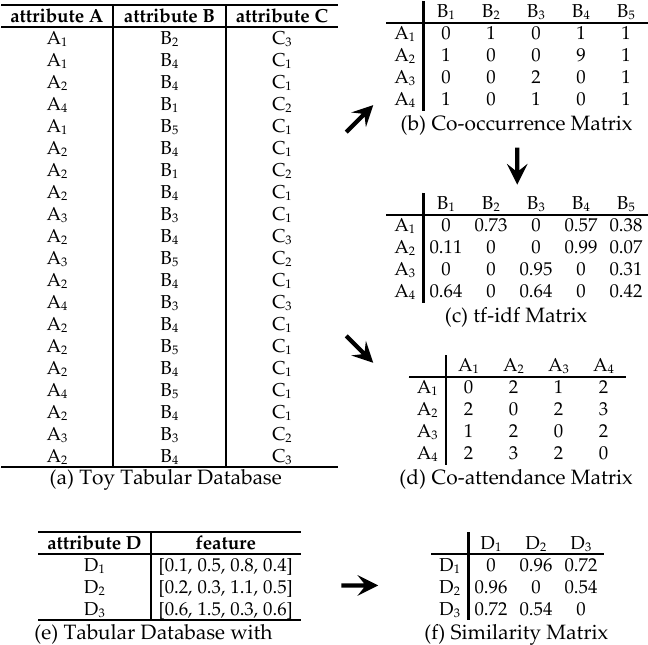}
    \caption{Relational database with derived matrices.}
    \label{tab_toydata}
\end{figure}

The preparation of entity-relation-matrices for text and graph are well studied in prior works~\cite{tang2015pte,pennington2014glove,tsitsulin2018verse}, so here we focus on describing the methods for a tabular database like the example shown in Fig.~\ref{tab_toydata}.a.
We consider different attributes (e.g., $A$, $B$, $C$) as different entity-types and each unique value in an attribute (e.g., $A_1$, $A_2$, $B_1$, $B_2$) as a different entity.
We refer to the entities stored in attribute $A$, $B$, and $C$ as type-$A$, type-$B$, and type-$C$ entities, respectively.

The most straightforward way of creating an entity-relation-matrix is to count the number of co-occurrences of two selected entities in the database.
For example, Fig.~\ref{tab_toydata}.b contains the number of times each pair of the type-$A$ and type-$B$ entity co-occur in the same row of Fig.~\ref{tab_toydata}.a.
If the objective is to analyze type-$A$ entities rather than type-$B$ entities, it is useful to apply the tf-idf technique to the entity-relation-matrix by considering type-$A$ entities as documents and type-$B$ entities as words.
Fig.~\ref{tab_toydata}.c shows the resulting matrix after applying tf-idf to the co-occurrence matrix.

Another way of deriving an entity-relation-matrix from Fig.~\ref{tab_toydata}.a is by evaluating the size of intersections between pairs of same type entities based on another entity-type.
We refer to the resulting entity-relation-matrix as a co-attendance matrix.
Continuing with the example in Fig.~\ref{tab_toydata}.a, consider if each type-$A$ entity corresponds to a person, and each type-$C$ entity corresponds to an event, then $A_1$ attends the events $C_1$ and $C_3$, while $A_2$ attends the events $C_1$, $C_2$ and $C_3$, respectively.
The intersection between $A_1$ and $A_2$ on attribute $C$ is $\{C_1, C_3\}$; therefore, the relationship strength is $2$.
Fig.~\ref{tab_toydata}.d shows the co-attendance matrix built using data shown in Fig.~\ref{tab_toydata}.a.
Such a method is a special case of the second-order proximity presented in~\cite{tang2015line}.
Note that while both the rows and columns of a co-attendance matrix correspond to entities of the same semantic type, they should still be treated as different entity-types in Algorithm~\ref{alg_mulmat}.
In such aspects, our framework is similar to the word2vec and GloVe approach~\cite{mikolov2013efficient,pennington2014glove}, where the target and the contextual words are considered as different entity-types even though the two represent the same set of words in a language.

In the case where each type-$D$ entity is associated with a set of numerical attribute data (Fig.~\ref{tab_toydata}.e), we can derive a pairwise similarity matrix for type-$D$ entities based on the cosine or other similarity measurement on numerical attributes as shown in Fig.~\ref{tab_toydata}.f.

\section{Cross-type Visualization}
\label{visual}
To qualitatively understand the association between different entities, it is important to visualize the entities of different entity-types in the same figure.
We first show that the straightforward approach of constructing a pairwise distance matrix without any further processing and projecting to a low-dimensional space using \textit{multi-dimensional scaling} (MDS)~\cite{kruskal1964multidimensional} or $t$-SNE~\cite{maaten2008visualizing} does not work for cross-type entity visualization.

\begin{figure}[ht]
    \centering
    \includegraphics[page=15, width=0.95\linewidth, trim={4.26in 2.69in 4.19in 2.58in},clip]{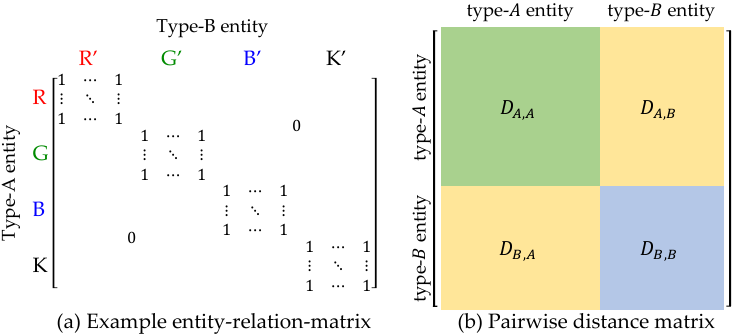}
    \caption{Visualization example.}
    \label{fig_vis_data}
\end{figure}

Consider a 20 $\times$ 20 entity-relation-matrix $M$ (Fig.~\ref{fig_vis_data}.a) that records the strength of co-occurrence relationship between type-$A$ and type-$B$ entities.
The diagonal of the matrix is occupied by four sub-matrices of size 5 $\times$ 5, each filled with ones, whereas the remaining elements in the matrix are all zeros.
Each row and column of the matrix can be considered as a feature vector for the type-$A$ and type-$B$ entities, respectively.
Consequently, the structure of the entity-relation-matrix suggests that both the type-$A$ and type-$B$ entities consist of four clusters.
We label the four clusters for type-$A$ entities as \{{\color{red}R}, {\color{green}G}, {\color{blue}B}, K\}, and as \{{\color{red}R'}, {\color{green}G'}, {\color{blue}B'}, K'\} for type-$B$ entities.
The sub-matrices of ones occupying the diagonal indicate that {\color{red}R} is strongly associated with {\color{red}R'}, {\color{green}G} is strongly associated with {\color{green}G'}, {\color{blue}B} is strongly associated with {\color{blue}B'}, and K is strongly associated with K'.
Therefore, a meaningful visualization should possess the following properties: (i) same type entities within the same cluster should be close to each other and (ii) cross-type embedding pairs strongly associated with each other should be close to each other.
For example, entities corresponding to the cluster {\color{red}R} should be close to \textit{each other} due to property (i), and to the entities corresponding to the cluster {\color{red}R'} due to property (ii).

\begin{figure}[ht]
    \centering
    \includegraphics[width=0.9\linewidth]{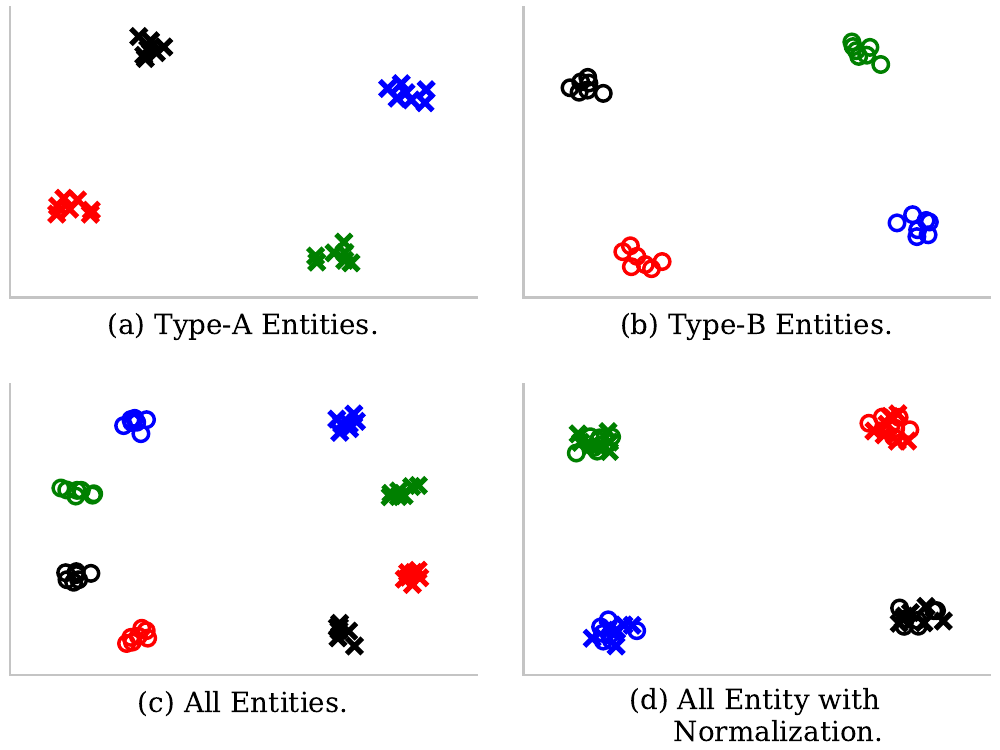}
    \caption{Normalization is required for proper visualization.}
    \label{fig_visual}
\end{figure}

We obtain the embeddings for type-$A$ and type-$B$ entities using the entity-relation-matrix $M$ as an input to Algorithm~\ref{alg_mulmat}.
With generated embeddings, the distance between a pair of embedding $v_i$ and $v_j$, either same type or cross-type, can be computed using $||v_i - v_j||_2$.
We organize the resulting pairwise distance matrix such that the distance between the same type entities is always placed along the diagonal, as shown in Fig.~\ref{fig_vis_data}.b.
As a result, the sub-matrices $D_{A,A}$ and $D_{B,B}$ store the distances between the same type entities whereas the sub-matrices $D_{A,B}$ and $D_{B,A}$ store the distances between cross-type entities with $D_{A,B} = D_{B,A}^\intercal$.
We generate a two-dimensional visualization using the pairwise distance matrix with \texttt{MDS}, as shown in Fig.~\ref{fig_visual}.
The marker's color corresponds to the cluster assignment, and the shape of the marker corresponds to the entity-type ($\times$ for type-$A$ entities and $\bigcirc$ for type-$B$ entities).

Fig.~\ref{fig_visual}.a and~\ref{fig_visual}.b shows the visualization for type-$A$ and type-$B$ entities when the pairwise distance matrix is computed only within the same type entities, i.e. using the sub-matrices $D_{A,A}$ and $D_{B,B}$, respectively.
We observe that, in compliance with property (i), the visualization consists of four well-separated clusters for both the type-$A$ and the type-$B$ entities.
However, when we compute the pairwise distance matrix using embeddings from both the type-$A$ and type-$B$ entities, the visualization does not comply with property (ii), as shown in Fig.~\ref{fig_visual}.c.
We observe that mirror images are formed and any association between cross-type entities is lost because the sub-matrices $D_{A,A}$, $D_{B,B}$ and $D_{A,B}$ are not commensurate with each other (note that $D_{B,A} = D_{A,B}^\intercal$).

To address the problem, we propose to normalize embeddings for type-$A$ entities and type-$B$ entities independently by removing the mean from each type's corresponding embeddings.
Let us say $\mathbf{V}_A$, and $\mathbf{V}_B$ contain the embeddings for type-$A$ and type-$B$ entities.
We first compute the mean for $\mathbf{V}_A$ $\mu_A = \frac{\sum_{v_i \in \mathbf{V}_A} v_i}{|\mathbf{V}_A|}$; then the normalized embeddings for type-$A$ can be computed with $v_i - \mu_A$ for each $v_i \in \mathbf{V}_A$.
We repeat the above steps on $\mathbf{V}_B$ to normalize $\mathbf{V}_B$.
The visualization now complies with both property (i) and (ii), as shown in Fig.~\ref{fig_visual}.d, where entities corresponding to the same color are close to each other regardless of the entity-types.
The normalization step is essential as the distance stored in each sub-matrix is not commensurable.
The example illustrated here provides an empirical validation of the proposed normalization method.
\section{Experiment}
\label{experiment}
In this section, we first demonstrate the effectiveness of the proposed framework on the restaurant embedding problem briefly described in Section~\ref{intro}.
Next, we evaluate our framework on the text- and graph-based data type using clustering metrics as a proxy for unsupervised representation problems.
We provide direct comparisons against the existing approaches, wherever possible, and empirically demonstrate the superior performance of the proposed framework.

\begin{table*}[t]
    \centering
    \caption{Top-5 restaurants retrieved using learned embeddings}
    \label{tab_cuisine}
    \resizebox{0.99\textwidth}{!}{
    \begin{tabular}{l|lllll}
    \toprule
    Cuisine Style   & 1               & 2                 & 3          & 4         & 5                \\ \hline \midrule
    Chinese         & Grand China Hing Chinese Restaurant & Sun China Chinese Restaurant          & China Golden Restaurant        & New Garden Chinese Restaurant & China One Chinese Restaurant          \\
    Greek           & Demetri's Greek Restaurant          & Niko's Gyro                           & Akropolis Greek Restaurant     & Akropolis Restaurant          & Zorbas Greek Restaurant               \\
    Indian/Pakistan & India's Tandoori                    & Maharaja Cuisine of India             & Paradise Cuisine of India      & Udipi Bhavan                  & The Everest Momo                      \\
    Italian         & Italian Garden Restaurant           & Benvenuto's Italian Grill             & Grazies Italian Grill          & Benvenuto's Italian Grill     & Benvenuto's Italian Grill             \\
    Japanese        & Asahi Japanese Restaurant           & \textcolor{red}{Daimo Chinese Restaurant}              & Ichi Sushi Grill House         & Japan Food Aki Restaurant     & Xushi Bento House                     \\
    Korean          & Woori Korean Restaurant             & Yuk Dae Jang                          & Yuk Dae Jang                   & Dan Moo Ji                    & Choon Chun Chicken BBQ                \\
    Mexican         & El Tenampa Mexican Restaurant       & Ajuua's Mexican Restaurant            & Javi's Mexican Food Restaurant & Mi Familia Mexican Restaurant & El Ranchito Mexican Restaurant        \\
    Pizza           & New York Pizza                      & Nunzio's Pizza and Italian Restaurant & Renna's Pizza                  & Milanese Brother's Pizza      & Padrino's Pizza and Family Restaurant \\
    Thai            & Thai Basil Restaurant               & Khun Suda Thai Cuisine                & Thai Erawan Restaurant         & Baan Khun Ya Thai Restaurant  & Royal Thai Restaurant                 \\
    Vietnamese      & Hai Duong Restaurant                & Bun Bo Hue Tay Do                     & Com Tam Thien Huong            & Binh Duong Quan               & Kim An 2 Vietnamese Cuisine       \\
    \bottomrule
    \end{tabular}
    }
\end{table*}

\subsection{Restaurant Embedding}
\label{exp_restaurant}

We qualitatively evaluate the proposed framework with a cuisine style-based restaurant retrieval task on the in-house credit card transaction database.
Cuisine style is one of the most important characteristics of a restaurant and plays a critical role in restaurant recommendation systems~\cite{yelp}.
However, in our credit card transaction database, it is quite common that the cuisine style information for a restaurant is either missing or incorrectly labeled.
Only 29\% of the restaurants are correctly labeled with a cuisine style.
We use a transaction database from July 2018 to December 2018 that consists of slightly more than 7 billion transactions between 2,947,930 users and 1,505,397 restaurants in the United States.
Our goal is to employ the proposed framework to learn restaurant embeddings that capture cuisine style information even if the cuisine style labels are missing or incorrect for a restaurant.
We extract five different entity-relation-matrices as described below from the database with entities like the restaurant, user, cuisine style, and words in a restaurant's name.
Each of these matrices contains incomplete/indirect evidence about a restaurant's cuisine style, albeit from different perspectives.

\begin{itemize}
    \item \textit{User-Restaurant} matrix records the number of times each user visited a particular restaurant.
    \item \textit{User-Cuisine style} matrix counts the number of times each user visited restaurants from different cuisine styles and could be used to identify a user's preferred taste.
    \item \textit{Restaurant-Cuisine style} matrix contains information about cuisine style for different restaurants.
    \item \textit{User-Word} matrix derives a distribution of words from the name of restaurants visited by each user.
    \item \textit{Restaurant-Word} matrix records the association between each restaurant and different words that occur in its name.
\end{itemize}

\begin{figure}[ht]
    \centering
    \includegraphics[width=0.90\linewidth]{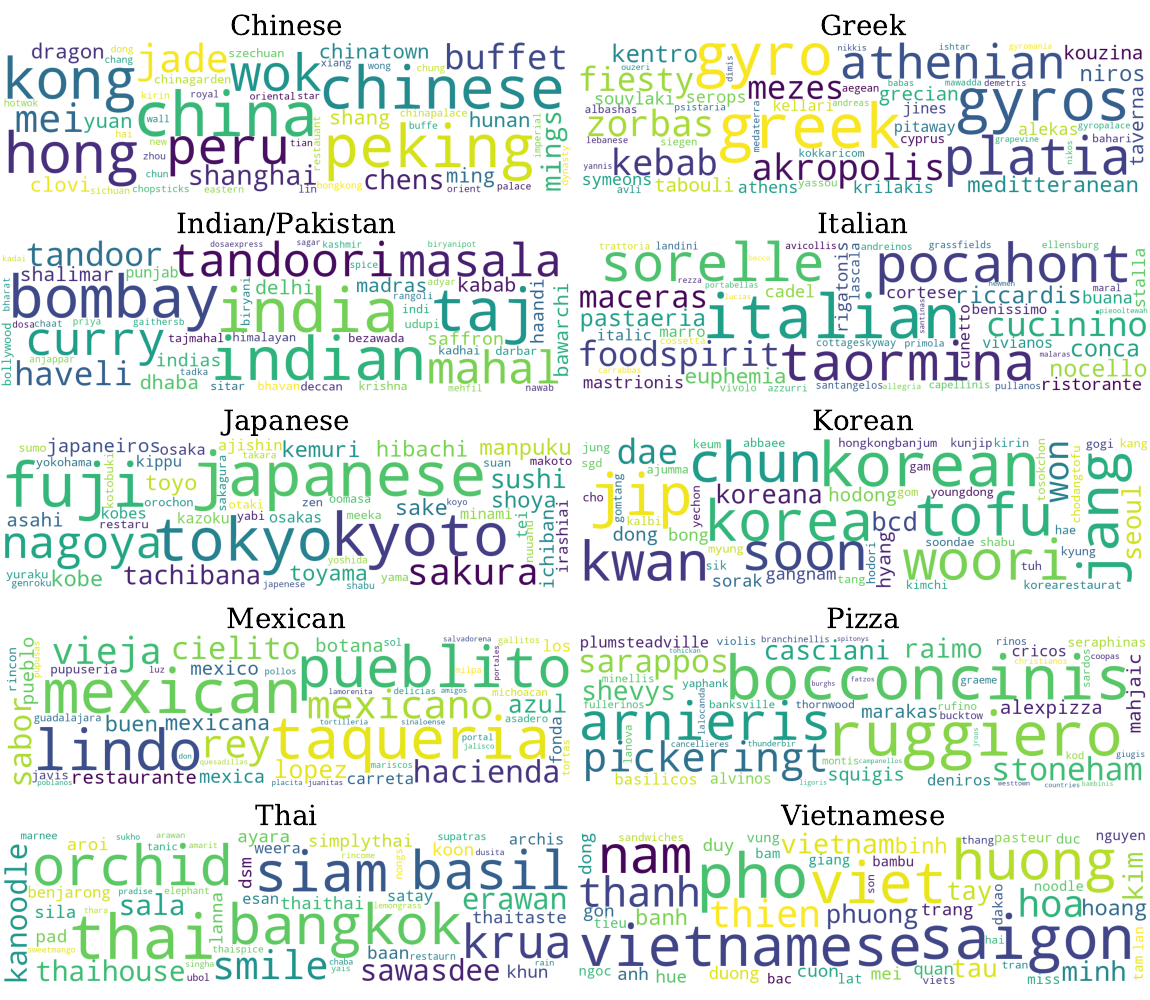}
    \caption{The word clouds generated for each cuisine style.
    }
    \label{fig_dnb_word}
\end{figure}

We first use the word cloud to highlight the association between cuisine style and word embeddings.
We choose ten cuisine styles and use the corresponding cuisine style embedding to find the 50 nearest word embeddings using cosine similarity.
The resulting word clouds are shown in Fig.~\ref{fig_dnb_word}.
The size of a word indicates the cosine similarity between the word embedding and the cuisine style embedding.
Based on visual inspection, the learned embeddings indeed capture common words used by restaurants of different cuisine styles.

We also repeat the above experiment on restaurant embeddings to find the top-5 restaurants that are closest to each of the ten cuisine style embedding as shown in TABLE~\ref{tab_cuisine}, and the nearest neighbor search amongst the restaurants \emph{without} a cuisine style label.
We assess the quality of the learned restaurant embeddings by manually checking the true cuisine style for each restaurant on the internet.
Only one out of fifty restaurants (highlighted in red) does not belong to the true cuisine style.
The average precision@5 for the embeddings generated by our framework is 98\%.
In comparison, the average precision@5 for randomly guessing the label would be 1.1\%, and if we learn the restaurant embeddings using Word2vec~\cite{mikolov2013efficient}, as suggested by~\cite{du2019pcard}, the performance increases only to 12\%.
The superior performance of our proposed framework can be attributed to the fact that the user is free to choose what information the embeddings should contain by supplying a relevant set of matrices to the framework.
On the contrary, the Word2vec~\cite{du2019pcard,mikolov2013efficient} approach fails to capture cuisine style information as the context for each restaurant mostly embeds the location information.

\subsection{Researcher Clustering}
\label{exp_network}
Next, we evaluate the proposed framework with researcher clustering task on an academic collaboration network.
\begin{itemize}
    \item \textit{AMiner Collaboration Network}~\cite{dong2017metapath2vec, tang2008arnetminer} dataset consists of 9,323,739 researchers and 3,194,405 papers from 3,883 Computer Science (CS) venues until 2016.
    We focus on 246,678 researchers that are labeled with exactly one of the eight CS research areas.
    \item \textit{DBLP Citation Network V10} (DBLP)~\cite{tang2008arnetminer, aminer2017dblpv10} dataset consists of citation relationship and supplementary information regarding 3,079,007 papers.
    We primarily use the DBLP dataset to extract the title and abstract, if available, for each paper in the AMiner dataset.
\end{itemize}

The current state-of-the-art approaches for researcher clustering
employ advanced heterogeneous graph embedding techniques, such as metapath2vec and metapath2vec++~\cite{dong2017metapath2vec}, where the embeddings are trained with sampled metapaths of the given heterogeneous graph.
To apply the proposed embedding framework, we derive the following entity-relation-matrices:
\begin{itemize}
    \item \textit{Researcher-Venue matrix} (RV), a co-occurrence matrix (Fig.~\ref{tab_toydata}.b) with attribute $A$ corresponding to researchers and attribute $B$ corresponding to venues.
    \item \textit{Researcher-Paper matrix} (RP), a co-occurrence matrix (Fig.~\ref{tab_toydata}.b) with attribute $A$ corresponding to researchers and attribute $B$ corresponding to papers.
    \item \textit{Paper-Venue matrix} (PV), a co-occurrence matrix (Fig.~\ref{tab_toydata}.b) with attribute $A$ corresponding to papers and attribute $B$ corresponding to venues.
    \item \textit{Co-appearance of Researchers matrix} (CR), a co-attendance matrix (Fig.~\ref{tab_toydata}.d) with attribute $A$ corresponding to researchers.
    \item \textit{Researcher Bag of Words matrix} (RBoW), a co-occurrence matrix (Fig.~\ref{tab_toydata}.b) built using the DBLP dataset with attribute $C$ corresponding to researchers and attribute $D$ corresponding to words.
    We only consider top 4,096 most frequent words to extract RBoW.
    \item \textit{Venue Bag of Words matrix} (VBoW), a co-occurrence matrix (Fig.~\ref{tab_toydata}.b) built using the DBLP dataset with attribute $C$ corresponding to venues and attribute $D$ corresponding to words.
    We only consider top 4,096 most frequent words to extract VBoW.
    \item \textit{Venue Semantics relationship matrix} (VS), a similarity matrix (Fig.~\ref{tab_toydata}.f) built using the DBLP dataset with attribute $C$ corresponding to venues.
    We use cosine similarity to compute the similarity between different venues with their corresponding BoW features.
\end{itemize}

We first apply $k$-means algorithm to the extracted researcher embeddings and then compare the clustering results with the ground-truth labels using \textit{normalized mutual information} (NMI)~\cite{cai2010locally} as evaluation metrics following~\cite{dong2017metapath2vec}.
The experimental results are summarized in TABLE~\ref{tab_aminer_per} with baseline graph embedding approaches like node2vec~\cite{grover2016node2vec}, LINE~\cite{tang2015line} and PTE~\cite{tang2015pte}.

\begin{table}[ht]
    \centering
    \caption{Experiment result on researcher clustering.}
    \label{tab_aminer_per}
    \resizebox{0.65\columnwidth}{!}{
    \begin{tabular}{l|c}
        \toprule
                                                                        & NMI    \\ \hline \midrule
        node2vec~\cite{grover2016node2vec} & 0.2941 \\
        LINE~\cite{tang2015line}                              & 0.6423 \\
        PTE~\cite{tang2015pte}                                          & 0.6483 \\
        metapath2vec~\cite{dong2017metapath2vec}                        & 0.7470 \\
        metapath2vec++~\cite{dong2017metapath2vec}                      & 0.7354 \\
        \hline
        RP                                                              & 0.0228 \\
        RBoW                                                            & 0.1107 \\
        RV                                                              & 0.3594 \\
        CR                                                              & 0.6948 \\
        RP+RBoW+RV+CR                                                   & 0.7372 \\
        RP+RBoW+RV+CR+VS                                                & 0.7478 \\
        RP+RBoW+RV+CR+VBoW                                              & 0.7499 \\
        RP+RBoW+RV+CR+PV                                                & 0.7553 \\
        RP+RBoW+RV+CR+VS+VBoW+PV                                        & \textbf{0.8562} \\
        \bottomrule
    \end{tabular}
    }
\end{table}

We apply our framework to various permutations of the entity-relation-matrices described above to understand the relative contribution of different matrices to the quality of embeddings.
First, we evaluate the performance of embeddings constructed using only one of the researcher-related matrices, such as RV, RP, CR, or RBoW.
We observe that the CR matrix contains the most relevant information for the researcher clustering task while the RP matrix contains the least relevant information.
Further, when all four matrices (RV, RP, CR, and RBoW) are used simultaneously, the resulting embedding outperforms the embedding learned with the CR matrix alone.
The improvement in the quality of embedding highlights the importance of learning from multiple entity-relation-matrices.
Next, we explore the possibility of incorporating more venue-related information.
We use the four researcher-related matrices (RV, RP, CR, and RBoW), plus one of the venue-related matrices (PV, VBoW, or VS) to extract the embedding.
Again, we find that the inclusion of a venue-related matrix, no matter whether it is PV, VBoW, or VS, enhances the performance of the embedding over the four-matrices setup.
Finally, when we use all of the seven entity-relation-matrices, our framework is capable of learning embeddings that outperform the current state-of-the-art graph embedding approaches: metapath2vec and metapath2vec++.

\begin{figure}[ht]
    \centering
    \includegraphics[page=3, width=0.99\linewidth, trim={4.165in 1.835in 4.165in 1.88in},clip]{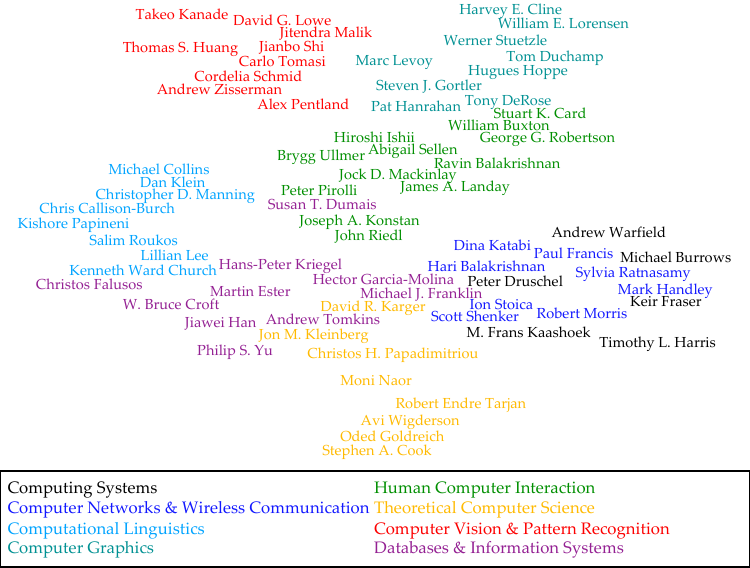}
    \caption{$t$\texttt{-SNE} 2-dimensional projection of a subset of researchers from each research area.}
    \label{fig_auth_auth}
\end{figure}

In addition to the quantitative evaluation, we also use the $t$-SNE visualization algorithm to assess if the learned embeddings are meaningful subjectively.
Fig.~\ref{fig_auth_auth} shows the two-dimensional projection of the learned embeddings.
Each researcher is color-coded with respect to the research area provided by the ground-truth label.

Overall, we observe that the researchers who work in the same research area are close to each other.
The relative location of different research areas also makes sense.
For example, ``Computer Vision \& Pattern Recognition'' is close to ``Computer Graphics'' as both are vision-related areas, and ``Computational Linguistics'' is next to ``Database \& Information System'' as both involve working with natural language data.
When we look closely at each researcher, we also find several interesting placements.
For example, Susan T. Dumais, labeled as ``Database \& Information System'' researcher, is placed among ``Human-Computer Interaction'' researchers because she also has several papers in ``Human-Computer Interaction.''
Researchers like David R Karger, Jon M. Kleinberg, Joseph A. Konstan, and John Riedl are placed at the boundary because of the interdisciplinary nature of their research.

\begin{figure}[ht]
    \centering
    \includegraphics[page=5, width=0.99\linewidth, trim={4.165in 1.79in 4.165in 1.79in},clip]{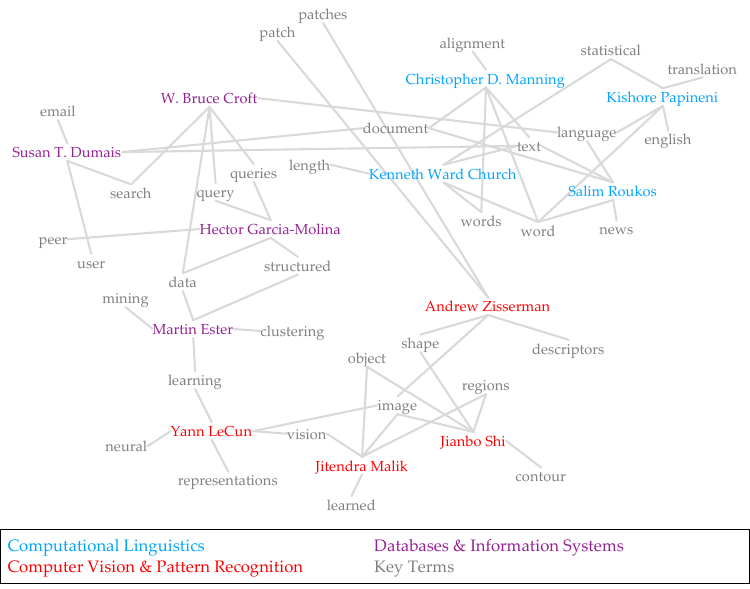}
    \caption{$t$\texttt{-SNE} 2-dimensional projection of a subset of researchers with relevant key words.}
    \label{fig_auth_word}
\end{figure}

We also select several researchers from ``Computer Vision \& Pattern Recognition,'' ``Computational Linguistics,'' and ``Database \& Information System,'' then plot them together with the top-five keywords associated with their research using the visualization method described in Section~\ref{visual} (Fig.~\ref{fig_auth_word}).
The association is determined by the cosine similarity between the embeddings of researchers and words.

We find that the position of researchers coincides with the position of relevant research areas.
The word ``data'' is close to researchers from ``Database \& Information System,'' ``text'' is close to researchers from ``Computational Linguistics,'' and ``image'' is close to researchers from ``Computer Vision \& Pattern Recognition.''
The keywords associated with each researcher also provides information regarding each researcher's focus in their respective research area.
For example, Yann LeCun, who is labeled with ``Computer Vision \& Pattern Recognition'' and has been working on neural network and representation learning, is highly related to the terms: ``neural" and ``representation.''

\begin{figure}[ht]
    \centering
    \includegraphics[width=0.99\linewidth]{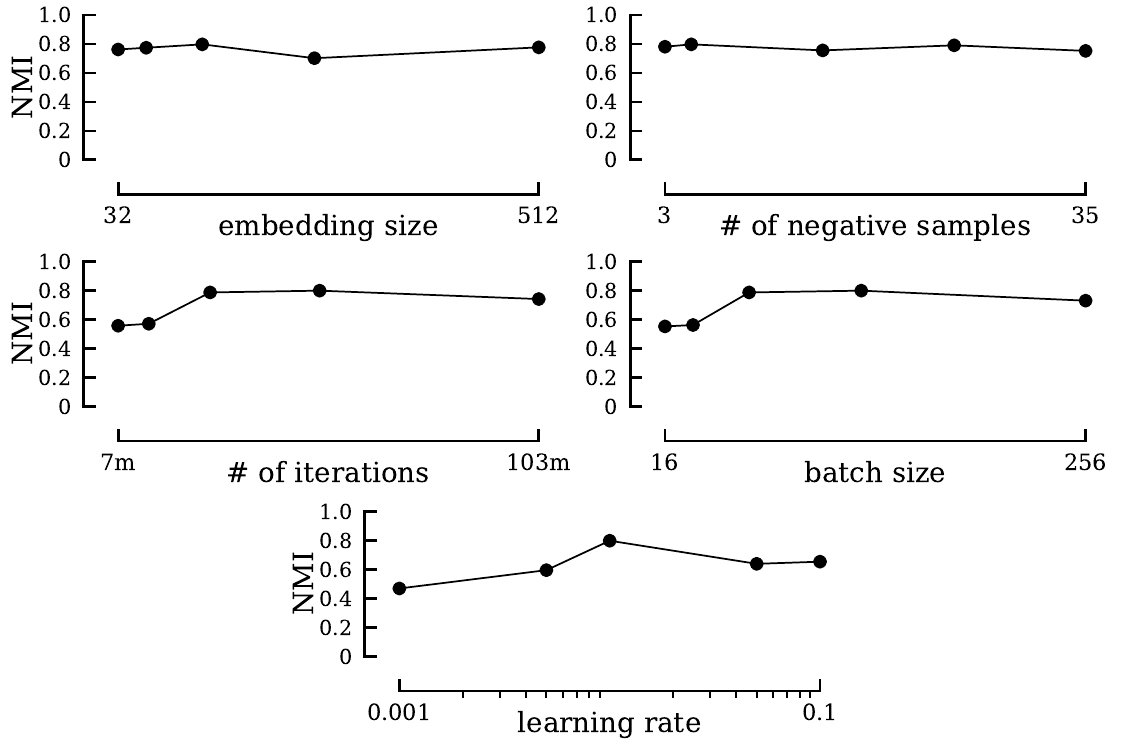}
    \caption{The researcher clustering performance as we vary each hyperparameter.
    }
    \label{fig_aminer_hyper}
\end{figure}

We further perform a hyperparameter sensitivity analysis using all the entity-relation-matrices, and the result is summarized in Fig.~\ref{fig_aminer_hyper}.
We find that our framework is not sensitive to the hyperparameters like embedding size and the number of negative samples.
However, hyperparameters that control the learning process, such as the number of iterations, batch size, and learning rate, do have a noticeable effect on the performance of resulting embeddings.
Such observation suggests that our framework could be improved if an advanced learning technique~\cite{daneshmand2016starting,chen2017outlier,smith2018disciplined} is adopted.

\subsection{Text Document Clustering}
\label{exp_text}
We also evaluate the proposed framework with document clustering task on the following text document datasets:
\begin{itemize}
    \item \textit{20 Newsgroups} (20news)~\cite{lang1995newsweeder, rennie2008newsgroups} dataset consists of 18,828 documents in raw text format from 20 different newsgroups.
    \item \textit{RCV1-v2} (RCV1)~\cite{lewis2004rcv1, lewis2004rcv1web} dataset consists of the bag-of-words feature of 804,414 documents.
    Each document is labeled with one or multiple news topics, such as economic performance, equity markets, etc.
    The total number of topics is 103.
    Following~\cite{yang2017towards}, we exclusively select 365,968 documents that are labeled with exactly one of the 20 most popular topics.
    \item \textit{The Westbury Lab Wikipedia} (wiki)~\cite{shaoul2010westbury} corpus consists of two million documents in raw text format obtained by crawling English articles from Wikipedia in April 2010.
\end{itemize}

The current state-of-the-art approach for text document clustering, \textit{deep clustering network} (DCN)~\cite{yang2017towards}, trains a deep learning model by jointly optimizing the self-reconstruction loss and clustering loss.
In addition to the DCN, we also compare the performance of the proposed framework against the following baseline approaches: Bag-of-Words (BoW), Term Frequency-Inverse Document Frequency (tf-idf), Word2vec (w2v), Doc2vec (d2v), and Pre-trained word2vec (pw2v).
For the proposed framework, we derive the following entity-relation-matrices:
\begin{itemize}
    \item \textit{BoW matrix} (BoW) that contains the BoW features for each document.
    The BoW matrix is a co-occurrence matrix (Fig.~\ref{tab_toydata}.b) with attribute $A$ corresponding to document and attribute $B$ corresponding to words.
    \item \textit{tf-idf matrix} (tf-idf) is simply the tf-idf normalized BoW matrix (Fig.~\ref{tab_toydata}.c) with attribute $A$ corresponding to documents and attribute $B$ corresponding to words.
    \item \textit{Word-Contextual word matrix} (WC) records the number of times a target word co-occurs with a contextual word within a fixed-size sliding window.
    The WC matrix is therefore an external co-occurrence matrix (Fig.~\ref{tab_toydata}.b) with attribute $A$ corresponding to target words and attribute $B$ corresponding to contextual words.
    We do not construct a WC matrix for the RCV1 dataset because the raw text format for the documents is not available.
\end{itemize}

For evaluation, we first apply $k$-means algorithm to the extracted document embeddings and then compare the clustering results with the ground-truth labels using \textit{normalized mutual information} (NMI)~\cite{cai2010locally}, \textit{adjusted Rand index} (ARI)~\cite{yeung2001details} and \textit{clustering accuracy} (ACC)~\cite{cai2010locally} as evaluation metrics following~\cite{yang2017towards}.
The experimental results are summarized in TABLE~\ref{tab_text}; we use \textit{EF} to denote the proposed embedding learning framework.

\begin{table}[ht]
    \centering
    \caption{Experiment result on text document clustering}
    \label{tab_text}
    \resizebox{0.90\columnwidth}{!}{
    \begin{tabular}{l|ccc|ccc}
        \toprule
                                   & \multicolumn{3}{c|}{20news} & \multicolumn{3}{c}{RCV1}                                                                 \\
                                   & NMI                         & ARI                         & ACC           & NMI           & ARI           & ACC           \\ \hline \midrule
        BoW                        & 0.04                        & 0.01                        & 0.09          & 0.31          & 0.06          & 0.31          \\
        tf-idf                     & 0.26                        & 0.09                        & 0.27          & 0.58          & 0.29          & 0.47          \\
        w2v                        & 0.18                        & 0.06                        & 0.18          & -             & -             & -             \\
        d2v                        & 0.11                        & 0.03                        & 0.14          & -             & -             & -             \\
        pw2v (google)              & 0.27                        & 0.12                        & 0.25          & 0.53          & 0.34          & 0.44          \\
        pw2v (wiki)                & 0.29                        & 0.14                        & 0.27          & 0.43          & 0.26          & 0.40          \\
        pw2v (giga)                & 0.31                        & 0.16                        & 0.29          & 0.47          & 0.33          & 0.47          \\
        DCN                      & 0.48                        & 0.34                        & 0.44          & 0.61          & 0.33          & 0.47          \\ \hline
        BoW/EF                     & 0.38                        & 0.16                        & 0.33          & 0.62          & 0.42          & 0.47          \\
        tf-idf/EF                  & 0.53                        & 0.37                        & 0.51          & 0.60          & 0.40          & 0.44          \\
        BoW/WC{\tiny 20news}/EF    & 0.50                        & 0.32                        & 0.50          & \textbf{0.63} & \textbf{0.49} & \textbf{0.54} \\
        tf-idf/WC{\tiny 20news}/EF & 0.54                        & 0.40                        & 0.56          & 0.62          & 0.42          & 0.48          \\
        BoW/WC{\tiny wiki}/EF      & 0.51                        & 0.32                        & 0.48          & 0.62          & 0.41          & 0.49          \\
        tf-idf/WC{\tiny wiki}/EF   & \textbf{0.56}               & \textbf{0.43}               & \textbf{0.61} & \textbf{0.63} & 0.44          & 0.48          \\
        \bottomrule
    \end{tabular}
    }
\end{table}

We observe that the baseline approach with tf-idf normalization outperforms BoW for both 20news and RCV1 dataset.
Interestingly, the pw2v approach that uses the average of word embeddings pre-trained with large-scale external datasets is an improvement over the tf-idf approach for the 20news dataset and comparable for the RCV1 dataset.
Further, the DCN approach, current state-of-the-art method for text document clustering, outperforms all the baseline approaches for both 20news and RCV1 dataset.

As discussed in Section~\ref{methodology}, our approach provides great flexibility in terms of the different permutations of entity-relation-matrices that could be used to extract embeddings.
First, we evaluate our embedding algorithm using BoW and tf-idf matrices, referred hereafter as BoW/EF and tf-idf/EF, respectively.
For the 20news dataset, we find that tf-idf/EF outperforms BoW/EF as well as the current state-of-the-art method (DCN).
However, the same observation does not hold for the RCV1 dataset: BoW/EF, tf-idf/EF, and DCN are comparable with each other, with BoW/EF marginally outperforming the other two.

Next, we explore the use of multiple entity-relation-matrices to improve clustering performance.
We apply our framework to a combination of the WC matrix and either the BoW matrix (BoW/WC/EF) or the tf-idf matrix (tf-idf/WC/EF).
We find that the addition of the WC matrix does improve the quality of learned embeddings as both BoW/WC{\tiny 20news}/EF and tf-idf/WC{\tiny 20news}/EF outperform BoW/EF and tf-idf/EF, respectively, for the 20news dataset.
Since the raw text format for the RCV1 dataset is not available, we cannot perform a similar experiment.
Instead, we draw some parallels for the RCV1 dataset by using an external dataset to construct a WC matrix.
To gauge the effect of simultaneously learning from multiple matrices, we simply adopt the WC{\tiny 20news} matrix for the RCV1 dataset and use it along with the available BoW or tf-idf matrix.
We find that the combination of BoW and WC{\tiny 20news} matrix, referred to as BoW/WC{\tiny 20news}/EF in TABLE \ref{tab_text}, outperforms the current state-of-the-art method (DCN) and the previous iterations of our approach (BoW/EF and tf-idf/EF) by a more substantial margin on two (out of three) evaluation metrics.

We further explore the use of external datasets in improving the quality of embeddings by comparing the performance of using WC{\tiny 20news} versus WC{\tiny wiki} on the 20news dataset.
We find that tf-idf/WC{\tiny wiki}/EF, a combination of WC{\tiny wiki} and tf-idf, outperforms the current state-of-the-art method (DCN) and any other permutations of the entity-relation-matrices on all three evaluation metrics.
For the RCV1 dataset, we find that neither BoW/WC{\tiny wiki}/EF nor tf-idf/WC{\tiny wiki}/EF further improves the embeddings in comparison to BoW/WC{\tiny 20news}/EF.
Overall, our framework is capable of learning useful embeddings from multiple entity-relation-matrices that could be extracted from internal or external datasets.

\begin{figure}[ht]
    \centering
    \includegraphics[width=0.90\linewidth]{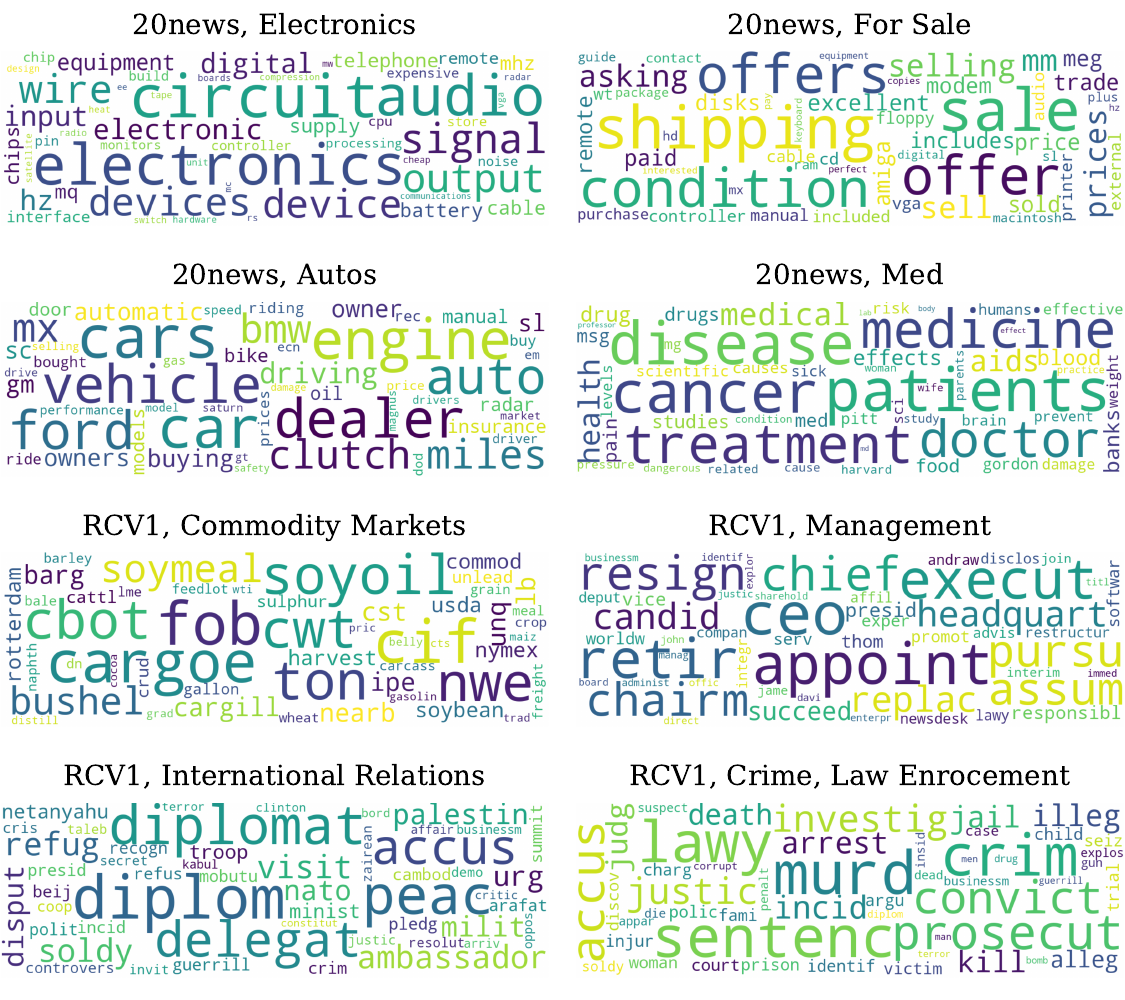}
    \caption{The word clouds generated from selected classes from 20news and RCV1.
    Misspelled words in RCV1's result are artifacts of the dataset.
    }
    \label{fig_word_cloud}
\end{figure}

A subjective analysis of the learned embeddings is performed.
Fig. \ref{fig_word_cloud} shows the word cloud generated from eight selected classes for both 20news and RCV1 dataset.
The size of the word indicates the average similarity between the word and each document.
From a visual inspection, it can be clearly seen that the document embeddings are more similar to embeddings of relevant words.

We also conduct a hyperparameters sensitivity analysis using the best performing setup for both datasets, and the conclusion is similar to the hyperparameters sensitivity analysis presented in Section.~\ref{exp_network}.



\section{Related Work}
\label{related}
The research on embedding learning methodology for words was popularized by the introduction of word2vec~\cite{mikolov2013efficient}, which learns word embeddings by modeling the co-occurrence of a word with its context.
The doc2vec algorithm~\cite{le2014distributed} improved upon word2vec by considering the document/paragraph-word relationship.
Levy and Goldberg~\cite{levy2014neural} analyzed the underlying algorithm of the word2vec approach, and showed that it implicitly performs matrix factorization.
Pennington et al.~\cite{pennington2014glove} proposed another word embedding learning algorithm, referred to as GloVe, that combined the benefits of both global matrix factorization~\cite{levy2014neural} as well as local context window methods~\cite{mikolov2013efficient}.
More recently, Arora et al.~\cite{arora2015rand} and Allen et al.~\cite{allen2019analogies} have continued to explore the theoretical justifications for the properties exhibited by embeddings learned through word2vec-based methods.
Further, PTE~\cite{tang2015pte} extended the original word2vec approach to a semi-supervised learning setting that results in a more substantial predictive power for a particular task, and ELMo~\cite{peters2018deep} generated context-aware word embeddings using bidirectional language models.
Since the primary goal for the embedding learning methods discussed above is to learn word embeddings from a text corpus, they cannot be directly applied to other data types.

Following the success of word embeddings in NLP, Perozzi et al.~\cite{perozzi2014deepwalk} proposed DeepWalk, an online learning algorithm for building graph embeddings.
DeepWalk learns the embedding in two stages: 1) performing a random walk on the input graph to sample pseudo sentences, and 2) applying word2vec to the sampled sentences.
The node2vec approach~\cite{grover2016node2vec} improved upon DeepWalk by using an alternative sampling strategy (i.e., biased random walk), which is more tailored towards the principles in network science.
metapath2vec~\cite{dong2017metapath2vec} further improved node2vec by adopting the concept of metapath in the sampling stage.
Line, proposed by \cite{tang2015line}, addressed other aspects of graph embedding learning, such as efficiency.
Recently, Qiu et al.~\cite{qiu2018network} proposed a general graph embedding learning framework by combining ideas from DeepWalk, node2vec, Line, and PTE.
The borrowing of ideas across different embedding learning methods is possible primarily because the underlying algorithm for each is still matrix factorization.
We also refer the reader to~\cite{cai2018comprehensive, goyal2018graph} for a more detailed survey about the development of graph embedding learning.
As before, all graph embedding learning methods discussed above also do not directly take a set of matrices as the input.

Apart from texts and graphs, effective embedding learning methods for other objects like behavior~\cite{wang2019tube} or stock~\cite{li2019individualized} have also been studied.
Nevertheless, to the best of our knowledge, this paper proposes the first embedding learning framework that achieves such a goal, is agnostic to the input data type and is highly flexible in terms of integrating domain knowledge.


\section{Conclusion}
\label{conclusion}
In this work, we propose a flexible embedding algorithm that is agnostic to the input data type.
We establish a more direct connection between the objective function optimized by the algorithm and the input, which makes it intuitive to inject domain knowledge into the embeddings.
We demonstrate that the proposed algorithm is highly flexible in terms of the possible methods for the construction of entity-relation-matrices.
By combining the relevant information from multiple entity-relation-matrices, the proposed algorithm outperforms the state-of-the-art approaches in various data mining tasks.
Finally, we also provide a visualization scheme to properly project the embeddings of different entity-types to 2- or 3-dimensional space.

\bibliographystyle{IEEEtran}



\end{document}